# ECG Arrhythmia Classification Using Transfer Learning from 2-Dimensional Deep CNN Features

Milad Salem, Shayan Taheri, Jiann-Shiun Yuan
Department of Electrical and Computer Engineering
University of Central Florida
Orlando, Florida, 32816 U.S.A.
Email: {miladsalem, shayan.taheri, jiann-shiun.yuan}@ucf.edu

*Abstract*— **Due to the recent advances in the area of deep learning, it has been demonstrated that a deep neural network, trained on a huge amount of data, can recognize cardiac arrhythmias better than cardiologists. Moreover, traditionally feature extraction was considered an integral part of ECG pattern recognition; however, recent findings have shown that deep neural networks can carry out the task of feature extraction directly from the data itself. In order to use deep neural networks for their accuracy and feature extraction, high volume of training data is required, which in the case of independent studies is not pragmatic. To arise to this challenge, in this work, the identification and classification of four ECG patterns are studied from a transfer learning perspective, transferring knowledge learned from the image classification domain to the ECG signal classification domain. It is demonstrated that feature maps learned in a deep neural network trained on great amounts of generic input images can be used as general descriptors for the ECG signal spectrograms and result in features that enable classification of arrhythmias. Overall, an accuracy of 97.23 percent is achieved in classifying near 7000 instances by ten-fold cross validation.**

## I. INTRODUCTION

Traditionally, identifying patterns in an electrocardiogram (ECG) signal and classifying the type of arrhythmia witnessed in the signal relied heavily on the features extracted from the signal. These features included statistical features, signal procession features, and medical features, which in the end required many engineering and optimization or domain expertise to deliver high accuracy [1]. Once the features were extracted, classification is traditionally done using support vector machine (SVM), Random Forest (RF), K-Nearest Neighbor (KNN), Feed Forward Neural Network, and a myriad of other classification tools that showed capability in this task [2].

Recent state-of-the-art performances achieved by deep learning methods in popular pattern recognition problems have motivated researchers and engineers to implement these techniques to the field of biomedical image and signal processing. In this regard, the deep learning methods have shown promising results in the ECG domain using Recurrent Neural Networks (RNN), specifically Long Short Term Memory Networks (LSTM) [2], and Convolutional Neural Network (CNN) [3]. One of the main advantages of using Deep Neural Networks (DNNs) is the fact that a neural network can automatically learn complex representative features directly from the data itself, therefore, eliminating the need for using manual feature extraction. Utilizing this merit gives an opportunity to create end-to-end learning systems that take ECG signals as input and output arrhythmia class prediction, while extracting the "deep features" autonomously. Another advantage of using a DNN is that deeper networks, in presence of sufficient amount of data, can deliver higher accuracy and better results in classification of fine-grained ECG signals.

In [3], it was shown that a 34-layered deep convolutional Neural Network can outperform board-certified cardiologist in detecting abnormalities and arrhythmias in ECG signal. In their work a dataset of near 64000 ECG recordings from approximately thirty thousand patients was used for training the DNN. However, this dataset is nearly 500 times larger than other datasets of its kind, demonstrating the fact that data volume was one of the main factors in achieving such performance.

In spite of the advantages that DNNs present in the ECG domain, one disadvantage is dominantly hindering the wide usage of these tools, the disadvantage of data volume. As compared to the traditional classification approaches DNNs require a considerable amount of data to be trained. This problem creates a gap between the dataset size and deep features, since the datasets that are publicly available in this domain lack in volume [3].

In order to fill this gap and arise to the problem of low ECG data volume versus high-performing deep features, in this work, we propose using transfer learning from the 2-dimensional domain. The image classification and object recognition domain are of the richest domains regarding the training data volume, in contrast to the ECG signal domain where datasets are relatively small. These domains contain sufficient amount of data to train DNNs and find feature maps that are capable of representing complex patterns in images. These learnt feature maps can be transferred to the ECG domain, if the 1-D ECG signal is transformed into a 2-D image using spectrograms. It will be shown that a Deep Neural Network, DenseNet, pre-trained on ImageNet classes can act thoroughly as feature extractor from ECG signal spectrograms on classification of four different rhythms including Normal Sinus Rhythm (Normal), Atrial Fibrillation and Flutter (AF), Ventricular Fibrillation (VF), and ST Segment Change (ST).

## II. BACKGROUND

### A. The Types of Arrhythmias Studied

A heart arrhythmia is a group of illness conditions in which the heart beats faster, slower, or in an irregular manner, commonly due to a disease. In this work three types of arrhythmias are of interest: (1) Ventricular Fibrillation, which is rapid and irregular electrical activity that causes the ventricles to fail to contract in a synchronized manner, resulting in cessation of cardiac output, (2) Atrial Fibrillation, which is defined as a very rapid and random disturbance of the atrium, (3) Change of the ST segment, including elevation or depression of the ST segment, based on the level of the segment is raised above or below the baseline.

### B. The Challenge of Fine-Grained Arrhythmia Classification

In independent studies, similar to this work, when the challenge of examining a unique abnormal rhythm is faced, the existence of sufficient amount of training data (containing records of the target rhythm) is vital. The details that need to be examined in an ECG signal are often fine-grained and similar, therefore, have patterns that are hard to detect, even for trained cardiologists [1]. However, the datasets existing in this domain contain a small amount of data or none for many abnormal rhythms. This challenge is amplified when one aims to approach it from a deep learning perspective; while deep learning can help the detection of the fine-grained patterns, training a deep neural network requires a humongous amount of data. If one attempts to train a deep neural network on a small amount of training data, overfitting may occur, which results in the classifier failing to detect patterns in unseen data. Solving this problem is crucial, since the trained classifier may be used for detection of ECG patterns obtained from human subjects, and failure to detect trained patterns on new unseen data may have "life or death" consequences.

Therefore, the availability of datasets, the amount of data, and reliability of the results of classifying unseen data propose a challenge in training deep neural networks for ECG pattern recognition. However, for the same reasons, we were encouraged to use deep learning from another perspective, which not only is resilient towards the aforementioned issues, but also helps resolve them. In this work, we examine the usage of "Transfer Learning" and "Off-the-Shelf CNN Features" for achieving high accuracy and reliability in abnormal rhythm detection in ECG signals. The achieved results is for the case of small training dataset.

### C. Transfer Learning and off-the-shelf CNN features

Knowledge learnt from pattern in one domain or task may be applicable to patterns in another domain or task. Utilization of Transfer Learning (TL) technique allows this knowledge to be transferred between two domains and used in the latter domain to enable classification. This technique is used and is sensible when there is lack of sufficient data, enough experience, and capable computing resources. An application of this technique is leveraging a pre-trained deep CNN for automatic feature extraction. The convolutional layers inside this network contain feature maps that are learnt during training on the original dataset and hold knowledge regarding the patterns existing in that dataset. These feature maps can act as feature extractor from another dataset. These "off-the-shelf" extracted features from the intermediate layers of a deep neural network are strong enough to conquer the hand-crafted features and be an ideal candidate for feature extraction [4].

In this work, we aim to bring the knowledge learnt from millions of images in the ImageNet dataset, via a deep neural network (DenseNet), to the ECG domain and classify a small dataset of thousands of instances using this neural network as a feature extractor. We show that the patterns learnt from the ImageNet dataset, which consists of many classes of images such as animals and objects, can be used to represent spectrograms of ECG signals.

*1) DenseNet*

Densely Connected Convolutional Neural Network (DenseNet) is a deep convolutional neural network with connections between each layer and every other layer in a feed-forward approach [5]. This type of connection helps attenuate the problem of vanishing-gradient, resulting in better training and feature propagation. DenseNet showed promising results in object recognition benchmark tasks. The architecture of DenseNet involves four Dense Blocks variable in length. In this work we use a pre-trained DenseNet-161, which has 161 convolutional layers through-out its structure and examine the outputs of these layers for feature extraction.

## III. METHODOLOGY

### A. Method Overview

In order to classify the input ECG signal into four classes of interest, the recordings are first cut, and the data is selected based on the annotations. Each data instance is then transformed into an image via using spectrograms. The images are then fed into a pre-trained DenseNet, a 161 layered deep CNN, and features are extracted by finding the output of 12 intermediate convolutional layers. In the end, these features are classified using SVM by ten-fold cross validation and the optimum layer is found based on their performance. The classification steps are shown in Fig. 1.

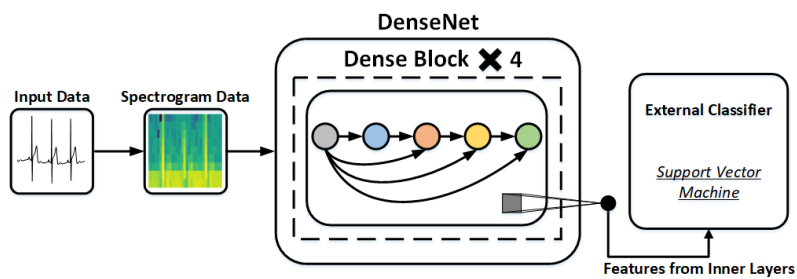

Fig. 1. Different stages of the classification system

*B. Data Sets in Use*

The data classified in this work is extracted from four different datasets: 1) MIT-BIH Atrial Fibrillation Dataset: This dataset consists of 24 recordings of nearly ten hours long, recorded from human subjects with atrial fibrillation (mostly paroxysmal). 2) The MIT-BIH Malignant Ventricular Arrhythmia Database: Containing 22 half-hour recordings, this dataset included patterns for ventricular tachycardia, ventricular flutter, and ventricular fibrillation. 3) European ST-T Database: including 90 annotated excerpts of ambulatory ECG recordings with different lengths, this dataset contains 367 episodes for ST change. 4) The MIT-BIH Normal Sinus Rhythm Database: this database includes 18 long-term ECG recordings of subjects that had no significant arrhythmias.

*C. Data Selection*

The aforementioned datasets contain many recordings and consequently many beats. In order to select data instances that hold enough information, we define a window size of 500 samples and cut each recording into data instances that contain this number of samples. This window size enables approximately 3 to 7 beats in each data instance.

The datasets in use contain arrhythmias that have been annotated by different cardiologists in different approaches. The AF and VF dataset have annotations marking the change in the type of the rhythm. For the ST dataset, the annotations mark the boundaries of each change in the ST level. In order to stay accurate, the data selection occurs near the annotation mark or near the peak of the change on all of the recordings in the dataset. In the Normal dataset, only the beats are annotated, therefore, the data selection occurs randomly around the annotated beats. In order to be diverse, data selections in all classes are done from all of the recordings in each dataset.

Overall, 7008 data instances are selected that show the symptoms and patterns of the rhythms of interest. This data volume is considerably small compared to the millions of data instances used in popular image classification tasks.

*D. Transformation from 1D Signal to 2D Image*

Since in this work a deep neural network pre-trained on images is used as feature extractor for ECG signals, the data instances need to be transformed into an image. In order to do so, we use spectrograms. Spectrograms are able to capture the changes in the power of the signal in an image by taking the Fourier transform of each partition of the signal. The number of partitions can be seen as a tune-able hyper-parameter which depends on the details of the signal and the relevant changes of the signal between classes. In this work 31 partitions were chosen to create the spectrograms. A sample of each class's spectrogram is shown in Fig. 2.

*E. Feature Representation Via DenseNet*

It has been shown that generic descriptors extracted from the inner layers of convolutional neural networks can be very potent in representing the input. It was shown in Nguyen et al. [6] that the layer of which the features have been extracted from can have direct effect on the accuracy of the recognition task.

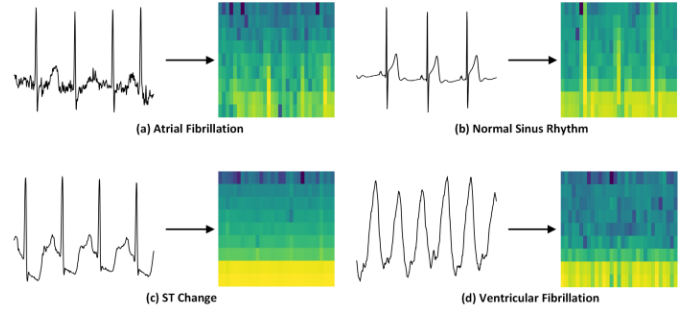

Fig. 2. Spectrogram of a sample data instance belonging to each class

This theory is intuitively understandable since each layer holds different feature maps which are activated in the presence of specific patterns. To perform feature extraction, the output of a few convolutional layers inside DenseNet are examined and considered as feature vectors. We randomly select twelve layers and extract a feature vector from each layer's output for each input image.

*F. Feature Selection and classification*

It is crucial to understand that each extracted feature vector contains many feature maps that exist on our pre-trained DenseNet and were formed to be able to distinguish between the thousand classes of ImageNet. Therefore, not all the patterns that exist on the feature maps of each layer may act as a good feature extractor. To increase the accuracy and to select these specific feature maps, we reduce the dimensionality of these vectors by selecting the feature maps that have high importance using the chi-squared test. The classification of each feature vector is done via a support vector classifier with a linear kernel. The classification is repeated with feature selection and the results are recorded.

IV. RESULTS

*A. Experimental Setup*

In order to extract the feature vectors from the spectrograms, we acquired a Keras implementation of DenseNet from [7]. The DenseNet-161 pre-trained weights and model were used, resulting in a deep neural network with 161 convolutional layers trained on a thousand classes of generic images. We examined 12 layers randomly and used the output of these layers as feature vectors. After feature extraction, the vectors were classified via linear SVM using ten-fold cross-validation. The process was repeated when feature selection was applied inside each fold.

*B. Results*

The result of feature extraction via DenseNet and classification via SVM are shown in Fig. 3. It is visible that some layers contain feature maps that perform well on the task of recognizing the patterns of the input data and result in higher accuracy. Moreover, the optimum layer for feature extraction is layer 112 both before and after feature selection is applied. Via selecting the features, the overall accuracy is generally increased nearly 2 percent, showing the importance of this task.

When examining different layers' performance on each class, it is clear that ST was the hardest class to classify in all layers, having the lowest F1 score regardless of the selected layer. Furthermore, this figure demonstrates that each layer has different capabilities in extracting features from different classes' inputs, having different patterns residing within.

Fig. 3. Results of SVM model on each layer's feature vectors, overall (left plot) and for each class (right plot).

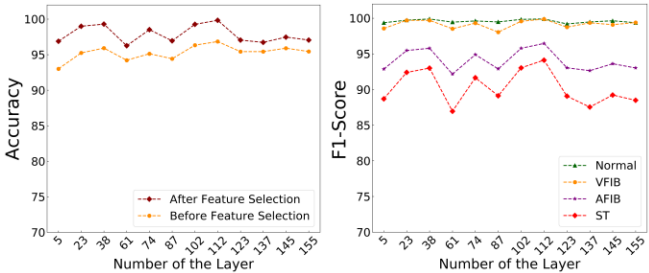

The best result for feature extraction via DenseNet is from the 112$^{th}$ layer, combined with classification via SVM and feature selection. This layer delivers 97.23 percent accuracy. The confusion matrix for the final result is shown in Fig. 4. Table 1 compares the results for different classes.

Fig. 4. The normalized confusion matrix of the best result achieved

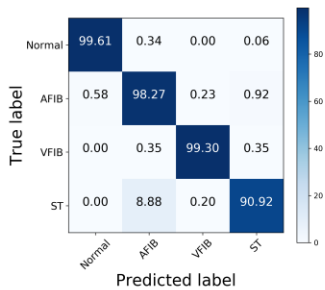

TABLE I. THE FINAL RESULT OF CLASSIFICATION FOR EACH CLASS USING DENSENET AS FEATURE EXTRACTOR.

| Class | Precision | Recall | F1 |
|---|---|---|---|
| Control | 99.17 | 99.61 | 99.39 |
| AF | 94.76 | 98.27 | 96.48 |
| VF | 99.22 | 99.30 | 99.26 |
| ST | 97.89 | 90.92 | 94.28 |

TABLE II. COMPARISON OF DIFFERENT CLASSIFICATION APPROACHES.

| Approach | #Classes | #Instances | Precision | Recall | Accuracy | F1 |
|---|---|---|---|---|---|---|
| 1D Signal+ SVM | 4 | 7008 | 57.46 | 56.66 | 61.00 | 48.27 |
| 2D Spec.+ SVM | 4 | 7008 | 89.43 | 88.33 | 89.04 | 88.85 |
| 2D Spec.+ DenseNet + SVM | 4 | 7008 | 97.76 | 97.02 | 97.23 | 97.35 |
| [3] | 14 | ~1,502,000 | 80.0 | 78.4 | -- | 77.6 |
| [8] | 5 | 100,389 | 63.5 | 60.3 | 97.6 | 61.85 |

In order to show the effects of bringing 1-D ECG signals to the 2-D domain using spectrograms, two additional SVM models were trained on raw 1-D signals and their spectrograms, respectively. The results of which are shown in Table II, alongside the results from using DenseNet as feature extractor, and two related works.

Table II demonstrates three points. Firstly, transforming ECG signals via spectrograms improves the accuracy of the model greatly. Secondly, utilization of DenseNet as feature extractor results in better discrimination of the spectrograms with furthered accuracy. Thirdly, our results have outperformed [8] and are arguably scalable to [3]. While [3] had a high number of classes with a higher possible overlap between them, the humongous size of the training data allowed having high precision and recall. On the other hand, [8] with a close number of classes to our work and more instances, achieved a lower F1-score. Therefore, via transfer learning and translation to 2-D domain, we were able to extract better representatives of ECG signals and improve classification results despite the small amount of data.

V. CONCLUSION

This work, to our understanding, is the first effort to use a deep CNN, pre-trained on millions of images, as a feature extractor from spectrograms of ECG signals. An accuracy of 97.23 percent is achieved in classifying a small dataset of approximately seven thousand samples of ECG signals, containing four classes of rhythms, using a pre-trained 161 layered DenseNet as a feature extractor and SVM for classification. The results show that firstly using spectrograms to transform ECG signals to images can preserve their fine-grained details; secondly feature maps learnt from a colossal amount of generic data in a deep neural network can act very well to represent spectrograms of ECG signals.